\documentclass[sigconf]{acmart}
\usepackage{balance}

\AtBeginDocument{%
  }

\setcopyright{acmlicensed}
\copyrightyear{2024}
\acmYear{2024}
\acmDOI{10.1145/3627673.3679891}

\acmConference[CIKM '24]{Proceedings of the 33rd ACM International Conference on Information and Knowledge Management}{October 21--25, 2024}{Boise, ID, USA}

\acmISBN{979-8-4007-0436-9/24/10}

\begin{document}

\title{COSCO: A Sharpness-Aware Training Framework for Few-shot Multivariate Time Series Classification}

\author{Jesus Barreda}
\affiliation{%
  \institution{University of Texas Rio Grande Valley}
  \city{Edinburg}
  \state{Texas}
  \country{USA}}
\email{jr.barreda1017@gmail.com}
\orcid{0009-0009-5613-6920}

\author{Ashley Gomez}
\affiliation{%
  \institution{University of Texas Rio Grande Valley}
  \city{Edinburg}
  \state{Texas}
  \country{USA}}
\email{ashley.gomez06@utrgv.edu}
\orcid{0009-0005-1367-4690}

\author{Ruben Puga}
\affiliation{%
  \institution{University of Texas Rio Grande Valley}
  \city{Edinburg}
  \state{Texas}
  \country{USA}}
\email{ruben.puga02@utrgv.edu}
\orcid{0009-0000-2865-977X}

\author{Kaixiong Zhou}
\affiliation{%
  \institution{North Carolina State University}
  \city{Raleigh}
  \state{North Carolina}
  \country{USA}}
\email{kzhou22@ncsu.edu}
\orcid{0000-0001-5226-8736}

\author{Li Zhang}
\affiliation{%
  \institution{University of Texas Rio Grande Valley}
  \city{Edinburg}
  \state{Texas}
  \country{USA}}
\email{li.zhang@utrgv.edu}
\orcid{0000-0003-3665-3989}

\renewcommand{\shortauthors}{Barreda et al.}

\begin{abstract}
  Multivariate time series classification is an important task with widespread domains of applications. Recently, deep neural networks (DNN) have achieved state-of-the-art performance in time series classification. However, they often require large expert-labeled training datasets which can be infeasible in practice. In few-shot settings, i.e. only a limited number of samples per class are available in training data, DNNs show a significant drop in testing accuracy and poor generalization ability. In this paper, we propose to address these problems from an optimization and a loss function perspective.  Specifically, we propose a new learning framework named \textbf{COSCO} consisting of a sharpness-aware minimization (SAM) optimization and a Prototypical loss function to improve the generalization ability of DNN for multivariate time series classification problems under few-shot setting. Our experiments demonstrate our proposed method outperforms the existing baseline methods. Our source code is available at: \url{https://github.com/JRB9/COSCO}.
\end{abstract}




\maketitle

\section{Introduction}
\begin{figure}
    \centering
    \resizebox{85mm}{!}{\includegraphics[width=90mm]{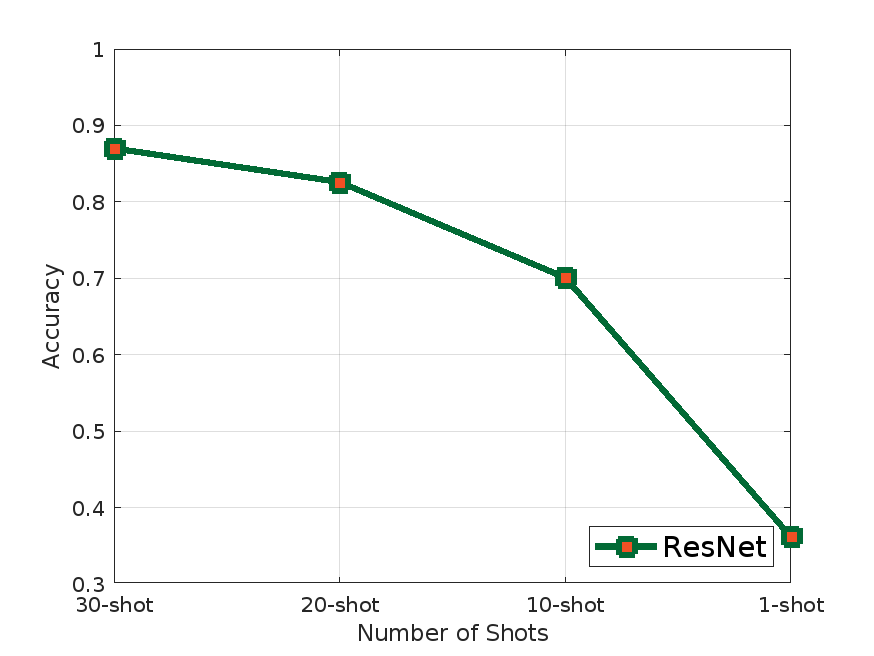}}
\caption{\fontseries{m}\selectfont Testing classification accuracy of the ResNet~\cite{he2016deep} model on the SpokenArabicDigits dataset. ResNet was trained and tested in 30-shot, 20, 10, and 1-shot settings.}
    \label{fig:intro}
\end{figure}

Multivariate time series classification has attracted significant research interest in the last two decades due to its wide applications~\cite{bagnall2018uea, zhang2020tapnet, ruiz2021great, karim2019multivariate, li2021shapenet, lai2023context}. Recently, deep neural networks (DNN) have surpassed classical distance-based methods (e.g. 1NN-DTW)~\cite{dempster2020rocket,ismail2019deep} and achieved state-of-the-art performance. However, their success often relies on large numbers of labels~\cite{finn2017model,zha2022towards,xi2023lb,xi2024efficient} from domain experts which are challenging to obtain in time series data~\cite{tong2022technology,zhang2022joint, wei2006semi}. Under a more realistic few-shot setting, where each class has only a limited number of expert-labeled training data~\cite{zha2022towards, xi2024efficient}, DNNs models show poor generalization ability to new data as illustrated by Figure~\ref{fig:intro}. For example, the testing accuracy of ResNet on the SpokenArabicDigits dataset from UCR Time Series Classification Archive drops sharply from 87.0\% under the 30-shot setting to 70.0\% under the 10-shot setting and falls dramatically to 36.2\% under the 1-shot setting.  
It has been noted~\cite{foret2020sharpness, miaorethinking, sun2024entropy} that commonly used loss functions such as cross-entropy, often exhibit a complex landscape and thus might not be not be sufficient to achieve satisfactory generalization~\cite{foret2020sharpness}. In particular, without sufficient labeled data or prior knowledge regularization, the large scale of trainable parameters often fall into sharp local minima during model training~\cite{du2021efficient, andriushchenko2022towards}, where the loss value increases rapidly with any small perturbation to the model weights. It has been observed that the sharp local minima are correlated with poor model generalization in various domains. In addition, time series signals are known to be very complex and noisy. Under few-shot learning setting, any outlier samples could severely affect the decision boundary learnt by the model and significantly hurt the model generalization~\cite{snell2017prototypical,ding2020graph}. 

To address this issue, we propose a new learning framework named \textbf{C}entroid \textbf{O}riented \textbf{S}harpness-\textbf{C}ontrolled \textbf{O}ptimization \\(\textbf{COSCO}) for few-shot multivariate time series classification. Specifically, COSCO proposes to incorporate sharpness-aware minimization (SAM)~\cite{foret2020sharpness} optimization technique with a perturb-then-update operation based on the neighborhood of loss values to combat complex landscapes in commonly used loss functions. In addition, COSCO utilizes a prototypical loss to replace the conventional cross-entropy loss to further improve the robustness against outlier samples in data scarcity situations. In summary, our main contributions is as follows: 1) propose a new learning framework designed for few-shot multivariate time series classification; 2) conduct comprehensive experimental results to demonstrate that COSCO outperforms baselines; 3) conduct ablation test to show the effectiveness of the modules. 

\section{Related Work} 
Multivariate time series classification has been one of the most extensively studied topics in the time series data mining domain over the past two decades~\cite{bagnall2018uea,zhang2020tapnet,liu2018time,karim2019multivariate}. Recent deep learning methods~\cite{he2016deep, zhang2020tapnet, dempster2020rocket, dempster2021minirocket} have achieved state-of-the-art performance, surpassing classical time series data mining methods such as 1NN-DTW~\cite{dau2019ucr}. However, most existing work relies on fully labeled datasets and requires extensive supervision. In few-shot settings, where only a limited amount of samples are available, their performance often experiences a significant decline. Existing research efforts have focused on semi-supervised learning and assume unlabeled training data are available. TapNet~\cite{zhang2020tapnet} utilizes a prototypical network to perform semi-supervised learning. SimTSC~\cite{zha2022towards} integrates pre-computed Dynamic Time Warping (DTW)~\cite{sakoe1971dynamic} and performs batch Graph Convolution Network (GCN)~\cite{kipf2016semi} to similarly perform semi-supervised learning. SimTSC requires pairwise DTW computation for all training, unlabeled, and testing data, which can be inefficient for large datasets. Meanwhile, LB-SimTSC~\cite{xi2023lb} uses LB-Keogh bound~\cite{keogh2005exact} instead of exact DTW distance, achieving comparable performance in semi-supervised learning. All above methods rely on the availability of unlabeled or testing data during training time, making them unsuitable for the more challenging situations encountered in few-shot time series classification problems. We differ from Tapnet~\cite{zhang2020tapnet} in two major ways. Firstly, our proposed prototypical loss is a loss function, rather than a network structure, and is specifically designed for few-shot learning without relying on testing data. In addition, Tapnet~\cite{zhang2020tapnet} is used in semi-supervised settings where test data is available during training. 
Recently, Sharpness-aware minimization techniques have demonstrated promising performance in domains such as image classification~\cite{foret2020sharpness, du2021efficient}, natural language processing~\cite{bahri2021sharpness}, and graph machine learning~\cite{wang2024efficient, xue2021cap, zhou2020towards, zhou2021dirichlet}. However, to the best of our knowledge, there is currently no research exploring how to improve the generalization ability of deep learning models through optimization techniques in multivariate time series classification problems. 

\section{Preliminaries} 
We begin by discussing the background of time series and then formulate the problem setting, followed by an introduction to deep learning models for time series. % 

\noindent\begin{definition}
    A \textbf{univariate time series} is denoted as $\mathbf{x} = [x_1, x_2, \cdots, x_T] \in \mathrm{R}^T$, where  $T$ is time length.  
    
\end{definition}
\noindent\begin{definition}
    A \textbf{multivariate time series} $\mathbf{X}$ consists of $M$ co-evolving univariate time series, i.e. $\mathbf{X} = [\mathbf{x}_1, \mathbf{x}_2, \cdots, \mathbf{x}_M]\in \mathrm{R}^{M\times T}$. 
\end{definition} 

\subsection{Problem Setting}
We formulate the multivariate time series classification (MTSC) problem. Let $\mathcal{X} = [\mathbf{X}_1, \mathbf{X}_2,\cdots, \mathbf{X}_{N}] \in \mathrm{R}^{N\times M\times T}$ denote a collection of multivariate time series (MTS), where $\mathbf{X}_i \in \mathrm{R}^{M\times T}$ is the $i$-th sample. Each instance $\mathbf{X}_i$ is associated with a label $y_i$ from classes $\mathcal{Y} \in \{1, \cdots, C\}$. Given a training dataset $\mathcal{X}^{\mathrm{train}} = [\mathbf{X}_1, \mathbf{X}_2, \cdots, \mathbf{X}_{N_t}]$ and the corresponding training labels $\mathcal{Y}^{\mathrm{train}} = [y_1, \cdots, y_{N_t}]$, the goal of MTSC problem is to learn a mapping function $f: \mathcal{X}\rightarrow\mathcal{Y}$ and apply it to predict the sample labels at testing set $\mathcal{X}^{\mathrm{test}}$. 

In this work, we consider a challenging setting of \textit{few-shot learning}, where the training data consists of a limited number of multivariate time series instances and labels. Particularly, there are only $k$ instances per class for training and $k$ is small. 

It is widely observed that deep learning models in various domains, such as computer vision and graph machine learning, are prone to converging to sharp local minima, where the loss value increases rapidly in the neighborhood around model weights~\cite{zhang2017mixup, cubuk2018autoaugment, foret2020sharpness, du2021efficient, andriushchenko2022towards}. The phenomenon of sharp local minima becomes more grievous in the few-shot setting, leading to poor generalization for classification for many test samples. However, no existing research discusses the connection between time-series deep learning models and their generalization performance under few-shot settings. We aim to bridge this gap by developing a model $f$ that can accurately predict on test data despite only training on a small number of samples. 

\section{Methods}
\subsection{Sharpness-aware Minimization} 
SAM~\cite{foret2020sharpness} has been recently proposed to smooth sharp local minima in large models and improve their generalization in realistic deployments. Specifically, SAM involves two successive steps to update the model: first, it generates perturbation gradients to adversarially shift the model weights to the neighbors associated with the worst loss; second, it explicitly updates the model to minimize this loss value. Through this perturb-then-update operation, SAM softens the loss landscape and improves generalization in domains such as visual and linguistic processing~\cite{chen2021vision}, graph data mining~\cite{wang2024efficient, xue2021cap}, 

Specifically, at each step, SAM trains deep neural networks by solving the
following min-max optimization problem:
\begin{equation}
\label{objective function}
\begin{aligned}
 \min_{\theta} \max_{\parallel\hat{\epsilon}\parallel_2\leq\rho} \mathcal{L}(\theta+\hat{\epsilon}),
\end{aligned}
\end{equation}
where $\theta$ denotes trainable parameters, $\hat\epsilon$ denotes perturbation gradient, $\rho$ is the maximum size of perturbation gradient, and $\mathcal{L}$ is the loss function such as cross-entropy loss for a classification problem. According to the above equation, SAM requires
two gradient computations at each training step: i) The inner maximization problem seeks to obtain the worst-case adversarial gradient, termed perturbing gradient. The perturbed model with $\hat\epsilon$ has the highest loss value centered around model weight $\theta$. ii) The outer minimization problem obtains an updating gradient used to finally improve the model. Via minimizing the worst-case loss, the deep learning model will converge to the optimal weights whose entire neighborhoods have lower loss values and lower curvature. In other words, the sharp local minima are flattened, which may result in enhanced generalization performance. 

In particular, considering training step $t+1$, SAM solves Problem~\ref{objective function} involving the following process:
\begin{equation}
\label{gradient1}
	\begin{aligned}
	\epsilon_t=\nabla_\theta\mathcal{L}(\theta_t), \quad &  \hat{\epsilon}_t = \rho \cdot \frac{\epsilon_t}{||\epsilon_t||_2}; \\
 \omega_t = \nabla_\theta\mathcal{L}(\theta_t+\hat{\epsilon}_t), \quad & \theta_{t+1} = \theta_t - \eta_t \cdot \omega_t. 
	\end{aligned}
\end{equation}
$\theta_t$ and $\theta_{t+1}$ denote the model weights at training steps $t$ and $t+1$, respectively. To obtain the intermediate perturbation gradient, SAM requires two complete forward and backward propagations, which obtain the desired generalization at the cost of time efficiency.

\subsection{Prototypical Loss} 
We introduce the prototypical loss used in COSCO. 
% intuition here
Instead of using commonly used cross-entropy and fully connected neural networks, we use prototypical loss which can integrate similarity between embeddings and utilize embedding centroid in each class to improve model generalizations. 

 Given a neural network $f: R^{M\times T} \rightarrow R^{E}$ where $E$ is the embedding size, Prototypical loss first computes an average embedding class centroid via:
 \begin{equation}
     \kappa_j = \frac{1}{|\mathcal{Y} = j|}\sum_{y_i=j} f(X_i),
 \end{equation}

\noindent where $|\mathcal{Y} = j|$ denotes total number of samples belong to class $j$.

Next, we compute the distance between each instance embedding and their corresponding class embedding centroid $\kappa_j$: 
\begin{equation}
    D_{i,j} = \Vert(f(X_i)-\kappa_j\Vert_2,
\end{equation}
\noindent where $\Vert \cdot \Vert_2$ denote $L_2$ norm. The prototypical loss is then defined as follows: 
\begin{equation}
    L(X_i, y_i) = - \frac{1}{N} \sum_{i=1}^{N} \log \left( \frac{\exp(-D_{i,j})}{\sum_{j=1}^N \exp(-D_{i,j})} \right)
\end{equation}

By using prototypical loss, we avoid the last fully connected neural network in typical neural networks to save the number of parameters and resist potential outlier instances in few-shot multivariate time series classification problems. 

\section{Experiment}
In this section, we will describe our experiment setting and evaluate the performance of the model. 
\subsection{Experiment Setting}

\subsubsection{Dataset} 
In our experiments, we use \textit{all} multivariate datasets that contain sufficient samples to be adapted into 10-shot settings (at least 10 samples per class) from the famous UEA time series classification repository~\cite{bagnall2018uea} to generate our few-shot data. 
We excluded datasets that took more than 4 hours to process. 
 
Table~\ref{tab:datasets_1} shows the information about the remaining 21 datasets we used in our experiment. 
\begin{table}[h]
\scalebox{0.7}{
\begin{tabular}{lllllll}
\hline
Dataset & 1-shot & 10-shot & Test Size & Length & \#Classes & Type\\
\hline
DuckDuckGeese & 5 &50 &50 & 236784 & 5 & AUDIO\\
Heartbeat & 2 &20& 205 & 405 & 2& AUDIO\\
JapaneseVowels & 9 &90& 370 & 29 & 9 & AUDIO\\
SpokenArabicDigits & 10 & 100& 2199 & 93 & 10 & AUDIO\\
FaceDetection & 2 &20& 3524 & 62 & 2 & EEG\\
FingerMovements & 2 &20& 100 & 50 & 2 & EEG\\
HandMovementDirection & 4&40 & 74 & 400 & 4 & EEG\\
MotorImagery & 2 &20& 100 & 3000 & 2 & EEG\\
SelfRegulationSCP1 & 2 &20& 293 & 896 & 2 & EEG\\
SelfRegulationSCP2 & 2 & 20&180 & 1152 & 2 & EEG\\
RacketSports & 4 & 40&152 & 30 & 4 & HAR \\
BasicMotions & 4 & 40 &40 & 100 & 4 & HAR\\
Epilepsy & 4 & 40 &138 & 207 & 4 & HAR\\
NATOPS & 6 &60& 180 & 51 & 6 & HAR\\
UWaveGestureLibrary & 8 & 80& 320 & 315 & 8 & HAR\\
ArticularyWordRecognition & 250 &25 & 300 & 144 & 25 & MOTION\\
CharacterTrajectories & 20 &200 & 1436 & 182 & 20 & MOTION\\
%EigenWorms & 128 & 131 & & 5 & 65& MOTION\\
PenDigits & 10 & 100& 3498 & 8 & 10 & MOTION\\
PEMS-SF & 7 & 70& 173 & 144 & 7 & OTHER\\
Libras & 15 &150& 180 & 45 & 15 & SENSOR\\
EthanolConcentration & 4 &40 & 263 & 1751 & 4 &SPECTRO\\

\hline
\end{tabular}}
\quad
\caption{\fontseries{m}\selectfont The dataset information of 1 and 10-shot few-shot data generated from UEA datasets~\cite{bagnall2018uea}. 1-shot and 10-shot columns show the few-shot training data size.}
\label{tab:datasets_1}
\end{table}

\begin{table*}[ht]
\centering
\scalebox{0.85}{
\begin{tabular}{llrrrrrrrrrr}
\hline
Datasets&& \multicolumn{5}{c}{1-shot}& \multicolumn{5}{c}{10-shot}\\ \cline{3-12} 
& Datatype & \multicolumn{1}{c}{1NN-DTW} & \multicolumn{1}{c}{1NN-ED} & \multicolumn{1}{c}{TapNet} & \multicolumn{1}{c}{ResNet} & \multicolumn{1}{c}{COSCO} & \multicolumn{1}{c}{1NN-DTW} & \multicolumn{1}{c}{1NN-ED} & \multicolumn{1}{c}{TapNet} & \multicolumn{1}{c}{ResNet} & \multicolumn{1}{c}{COSCO} \\ \hline
DuckDuckGeese& Audio& 0.280& 0.200& 0.220& \textbf{0.316}& 0.268& 0.520& 0.500& 0.212& \textbf{0.612}& 0.580\\
Heartbeat& Audio& 0.727& 0.712& 0.707& 0.649& \textbf{0.736}& 0.576& \textbf{0.620}& 0.543& 0.607& 0.619\\
JapaneseVowels& Audio& 0.776& 0.741& 0.465& \textbf{0.783}& 0.724& \textbf{0.962}& 0.943& 0.815& 0.900& 0.910\\
SpokenArabicDigits& Audio& \textbf{0.545}& 0.467& 0.123& 0.327& 0.352& \textbf{0.879}& 0.840& 0.739& 0.710& 0.824\\
FaceDetection& EEG& \textbf{0.500}& 0.497& 0.497& 0.496& \textbf{0.500}& \textbf{0.509}& 0.507& 0.500& 0.500& 0.502\\
FingerMovements& EEG& 0.430& 0.410& \textbf{0.508}& 0.442& 0.466& 0.490& 0.460& 0.496& 0.470& \textbf{0.506}\\
HandMovementDirection& EEG& 0.243& 0.162& 0.211& 0.246& \textbf{0.284}& 0.216& 0.243& 0.327& \textbf{0.343}& 0.314\\
MotorImagery& EEG& 0.500& \textbf{0.510}& \textbf{0.510}& 0.424& 0.424& 0.430& 0.450& \textbf{0.486}& 0.472& 0.458\\
SelfRegulationSCP1& EEG& 0.478& 0.478& 0.439& 0.657& \textbf{0.661}& \textbf{0.707}& 0.693& 0.617& 0.671& 0.651\\
SelfRegulationSCP2& EEG& 0.467& 0.478& 0.512& \textbf{0.546}& 0.520& \textbf{0.561}& 0.544& 0.507& 0.466& 0.464\\
BasicMotions& HAR& 0.750& 0.425& 0.410& \textbf{0.950}& 0.900& 0.975& 0.600& 0.675& \textbf{1.000}& \textbf{1.000}\\
Epilepsy& HAR& 0.399& 0.261& 0.254& \textbf{0.622}& 0.564& 0.877& 0.551& 0.422& 0.926& \textbf{0.933}\\
NATOPS& HAR& 0.522& 0.406& 0.227& \textbf{0.608}& 0.591& 0.822& 0.767& 0.556& 0.826& \textbf{0.848}\\
RacketSports& HAR& 0.375& 0.355& 0.346& 0.541& \textbf{0.554}& 0.803& 0.691& 0.576& \textbf{0.849}& 0.842\\
UWaveGestureLibrary& HAR& \textbf{0.728}& 0.616& 0.469& 0.397& 0.395& \textbf{0.909}& 0.816& 0.709& 0.767& 0.769\\
ArticularyWordRecognition & Motion& \textbf{0.870}& 0.787& 0.457& 0.715& 0.698& \textbf{0.983}& 0.970& 0.770& 0.970& 0.955\\
CharacterTrajectories& Motion& \textbf{0.783}& 0.569& 0.345& 0.699& 0.715& 0.954& 0.906& 0.780& 0.964& \textbf{0.975}\\
PenDigits& Motion& 0.617& 0.617& 0.365& \textbf{0.619}& \textbf{0.619}& 0.867& 0.867& 0.790& 0.895& \textbf{0.903}\\
PEMS-SF& Other& \textbf{0.318}& \textbf{0.318}& 0.289& 0.208& 0.305& 0.462& 0.515& 0.571& 0.490& \textbf{0.638}\\
Libras& Sensor& 0.472& 0.422& 0.260& \textbf{0.549}& 0.529& 0.850& 0.806& 0.610& \textbf{0.884}& 0.871\\
EthanolConcentration& Spectro& 0.278& \textbf{0.316}& 0.244& 0.261& 0.262& 0.251& \textbf{0.300}& 0.268& 0.267& 0.243\\ \hline
Average Accuracy ($\uparrow$) && \textbf{0.527}& 0.464& 0.374& 0.525& \textbf{0.527}& 0.695& 0.647& 0.570& 0.695& \textbf{0.705}\\
Average Rank ($\downarrow$) && 2.429& 3.190& 4.190& 2.619& \textbf{2.333}& 2.714& 3.048& 4.048& 2.619& \textbf{2.429}\\ \hline
\quad
\end{tabular}}
\caption{\fontseries{m}\selectfont Testing accuracy results for baselines compared to our proposed COSCO method. Best accuracy (higher is better) and best average rank (lower is better) are in bold. }
\label{tab:proto_results}

\end{table*}

To create few-shot datasets, we randomly select $k$ instances from each class in the datasets to form the training sets. We test with $k=1$ and $k=10$ for all methods. Each experiment was repeated 5 times for each dataset, and the average testing classification accuracy over the 5 trials was reported. Note that for fair comparison, the training and testing data were normalized in all experiments, except for audio data -- which is already within the appropriate range, thus no need for additional normalization.
 
\subsection{Baseline models} \quad
For COSCO, we use the same ResNet architecture from~\cite{zha2022towards} as our backbone. We set the neighborhood parameter $\rho = 0.1$ for SAM across all 1-shot and 10-shot data in this experiment. We use ResNet as one of the baselines. To maintain consistency, we apply the same learning rate and train for 100 epochs as with the ResNet baseline. 
We use the implementation from AEON\footnote{https://www.aeon-toolkit.org/} for 1-nearest-neighbor with Dynamic Time Warping (1NN-DTW), Euclidean distance methods (1NN-ED), and TapNet~\cite{zhang2020tapnet}. For 1NN-DTW, we set the warp window size to 0.1 following~\cite{dau2019ucr}. For TapNet, we use the default parameters and train for 100 epochs, consistent with our ResNet configuration. TapNet utilizes the Adam optimizer with a learning rate of 0.01. 

For both ResNet~\cite{he2016deep} and our COSCO models, we use the architecture from ~\cite{zha2022towards} with Stochastic Gradient Descent (SGD) optimizer as our backbone models. We set the SGD parameters with a momentum of 0.9 and a learning rate of 0.01, and train for 100 epochs throughout our experiments.  
The experiments are conducted on Google Colab\footnote{https://colab.research.google.com/} using a T4 16GB GPU.

\subsection{Experiment 1: Compare with Baselines}  
We first conduct experiments with baseline methods. 
Table~\ref{tab:proto_results} shows the testing accuracy for all methods. For accuracy, COSCO ties the highest average performance in 1-shot setting with 1NN-DTW, and achieves the highest average accuracy in 10-shot setting. For ranking, COSCO achieves the best average rank compared with all other methods. Compare with ResNet, COSCO consistently performed better than its backbone ResNet in both average accuracy and ranking, showing the effectiveness of our proposed learning framework.  
COSCO achieved the highest accuracy in the PEMS-SF 10-shot dataset, surpassing the second-place model by over 6$\%$. In the RacketSports dataset (HAR), COSCO excelled in the 1-shot setting, outperforming 1NN-DTW by over 17$\%$. The experiments demonstrate the ability of COSCO to handle a diverse category of data while improving average performance in few-shot settings.

\begin{table}[hh]
\scalebox{1}{
\begin{tabular}{lrrr}
\hline
Average Rank ($\downarrow$)  & \multicolumn{1}{l}{1-shot} & \multicolumn{1}{l}{10-shot} & \multicolumn{1}{l}{Overall} \\ \hline
COSCO & 2.000& \textbf{1.762}& \textbf{1.881}\\
COSCO w/o PL   & \textbf{1.810}& 2.048& 1.929\\
COSCO w/o (SAM \& PL) & 2.048& 2.000& 2.024\\
 \hline
\end{tabular}}
\caption{\fontseries{m}\selectfont Average Rank results for the COSCO method, compared against COSCO without (w/o) prototypical loss (PL), and COSCO without SAM and Prototypical Loss (SAM \& PL). A lower rank indicates better performance, with the best rank shown in bold.}
\label{tab:ablation_results}
\end{table}

\subsection{Experiment 2: Ablation Test of COSCO} 
We conduct an ablation test to demonstrate the effectiveness of both the Prototypical Loss function and SAM optimization technique. We step-wisely remove prototypical loss (PL) and SAM optimization and report the average rank performance on all the 1-shot and 10-shot datasets, as shown in  Table~\ref{tab:ablation_results}. In the 1-shot data setting, COSCO surpasses vanilla ResNet but falls short against SAM. In the 10-shot setting, COSCO consistently achieves 1st place (1.762) across all 21 datasets. Overall, COSCO shows the best performance out of three (1.881) and the vanilla ResNet model, which is COSCO without SAM and PL performs the worst (2.024). This shows our proposed modules are effective in general. 

\section{Conclusion}
In this paper, we propose to address these problems from an optimization and a loss function perspective.  Specifically, we propose a new learning framework named \textbf{COSCO} consisting of a sharpness-aware minimization (SAM) optimization and a Prototypical loss function to improve the generalization ability of DNN for multivariate time series classification problems under few-shot learning setting. Our experiments demonstrate our proposed method outperforms the existing baseline methods.
\newpage 
\bibliographystyle{ACM-Reference-Format}
\balance
\bibliography{classification}


\begin{thebibliography}{37}

%%% the Web address.

\ifx \showCODEN    \undefined \def \showCODEN     #1{\unskip}     \fi
\ifx \showDOI      \undefined \def \showDOI       #1{#1}\fi
\ifx \showISBNx    \undefined \def \showISBNx     #1{\unskip}     \fi
\ifx \showISBNxiii \undefined \def \showISBNxiii  #1{\unskip}     \fi
\ifx \showISSN     \undefined \def \showISSN      #1{\unskip}     \fi
\ifx \showLCCN     \undefined \def \showLCCN      #1{\unskip}     \fi
\ifx \shownote     \undefined \def \shownote      #1{#1}          \fi
\ifx \showarticletitle \undefined \def \showarticletitle #1{#1}   \fi
\ifx \showURL      \undefined \def \showURL       {\relax}        \fi
% invisible to TeX
\providecommand\bibfield[2]{#2}
\providecommand\bibinfo[2]{#2}
\providecommand\natexlab[1]{#1}
\providecommand\showeprint[2][]{arXiv:#2}

\bibitem[Andriushchenko and Flammarion(2022)]%
        {andriushchenko2022towards}
\bibfield{author}{\bibinfo{person}{Maksym Andriushchenko} {and} \bibinfo{person}{Nicolas Flammarion}.} \bibinfo{year}{2022}\natexlab{}.
\newblock \showarticletitle{Towards understanding sharpness-aware minimization}. In \bibinfo{booktitle}{\emph{International Conference on Machine Learning}}. PMLR, \bibinfo{pages}{639--668}.
\newblock


\bibitem[Bagnall et~al\mbox{.}(2018)]%
        {bagnall2018uea}
\bibfield{author}{\bibinfo{person}{Anthony Bagnall}, \bibinfo{person}{Hoang~Anh Dau}, \bibinfo{person}{Jason Lines}, \bibinfo{person}{Michael Flynn}, \bibinfo{person}{James Large}, \bibinfo{person}{Aaron Bostrom}, \bibinfo{person}{Paul Southam}, {and} \bibinfo{person}{Eamonn Keogh}.} \bibinfo{year}{2018}\natexlab{}.
\newblock \showarticletitle{The UEA multivariate time series classification archive, 2018}.
\newblock \bibinfo{journal}{\emph{arXiv preprint arXiv:1811.00075}} (\bibinfo{year}{2018}).
\newblock


\bibitem[Bahri et~al\mbox{.}(2021)]%
        {bahri2021sharpness}
\bibfield{author}{\bibinfo{person}{Dara Bahri}, \bibinfo{person}{Hossein Mobahi}, {and} \bibinfo{person}{Yi Tay}.} \bibinfo{year}{2021}\natexlab{}.
\newblock \showarticletitle{Sharpness-aware minimization improves language model generalization}.
\newblock \bibinfo{journal}{\emph{arXiv preprint arXiv:2110.08529}} (\bibinfo{year}{2021}).
\newblock


\bibitem[Chen et~al\mbox{.}(2021)]%
        {chen2021vision}
\bibfield{author}{\bibinfo{person}{Xiangning Chen}, \bibinfo{person}{Cho-Jui Hsieh}, {and} \bibinfo{person}{Boqing Gong}.} \bibinfo{year}{2021}\natexlab{}.
\newblock \showarticletitle{When vision transformers outperform resnets without pre-training or strong data augmentations}.
\newblock \bibinfo{journal}{\emph{arXiv preprint arXiv:2106.01548}} (\bibinfo{year}{2021}).
\newblock


\bibitem[Cubuk et~al\mbox{.}(2018)]%
        {cubuk2018autoaugment}
\bibfield{author}{\bibinfo{person}{Ekin~D Cubuk}, \bibinfo{person}{Barret Zoph}, \bibinfo{person}{Dandelion Mane}, \bibinfo{person}{Vijay Vasudevan}, {and} \bibinfo{person}{Quoc~V Le}.} \bibinfo{year}{2018}\natexlab{}.
\newblock \showarticletitle{Autoaugment: Learning augmentation policies from data}.
\newblock \bibinfo{journal}{\emph{arXiv preprint arXiv:1805.09501}} (\bibinfo{year}{2018}).
\newblock


\bibitem[Dau et~al\mbox{.}(2019)]%
        {dau2019ucr}
\bibfield{author}{\bibinfo{person}{Hoang~Anh Dau}, \bibinfo{person}{Anthony Bagnall}, \bibinfo{person}{Kaveh Kamgar}, \bibinfo{person}{Chin-Chia~Michael Yeh}, \bibinfo{person}{Yan Zhu}, \bibinfo{person}{Shaghayegh Gharghabi}, \bibinfo{person}{Chotirat~Ann Ratanamahatana}, {and} \bibinfo{person}{Eamonn Keogh}.} \bibinfo{year}{2019}\natexlab{}.
\newblock \showarticletitle{The UCR time series archive}.
\newblock \bibinfo{journal}{\emph{IEEE/CAA Journal of Automatica Sinica}} \bibinfo{volume}{6}, \bibinfo{number}{6} (\bibinfo{year}{2019}), \bibinfo{pages}{1293--1305}.
\newblock


\bibitem[Dempster et~al\mbox{.}(2020)]%
        {dempster2020rocket}
\bibfield{author}{\bibinfo{person}{Angus Dempster}, \bibinfo{person}{Fran{\c{c}}ois Petitjean}, {and} \bibinfo{person}{Geoffrey~I Webb}.} \bibinfo{year}{2020}\natexlab{}.
\newblock \showarticletitle{ROCKET: exceptionally fast and accurate time series classification using random convolutional kernels}.
\newblock \bibinfo{journal}{\emph{Data Mining and Knowledge Discovery}} \bibinfo{volume}{34}, \bibinfo{number}{5} (\bibinfo{year}{2020}), \bibinfo{pages}{1454--1495}.
\newblock


\bibitem[Dempster et~al\mbox{.}(2021)]%
        {dempster2021minirocket}
\bibfield{author}{\bibinfo{person}{Angus Dempster}, \bibinfo{person}{Daniel~F Schmidt}, {and} \bibinfo{person}{Geoffrey~I Webb}.} \bibinfo{year}{2021}\natexlab{}.
\newblock \showarticletitle{Minirocket: A very fast (almost) deterministic transform for time series classification}. In \bibinfo{booktitle}{\emph{Proceedings of the 27th ACM SIGKDD Conference on Knowledge Discovery \& Data Mining}}. \bibinfo{pages}{248--257}.
\newblock


\bibitem[Ding et~al\mbox{.}(2020)]%
        {ding2020graph}
\bibfield{author}{\bibinfo{person}{Kaize Ding}, \bibinfo{person}{Jianling Wang}, \bibinfo{person}{Jundong Li}, \bibinfo{person}{Kai Shu}, \bibinfo{person}{Chenghao Liu}, {and} \bibinfo{person}{Huan Liu}.} \bibinfo{year}{2020}\natexlab{}.
\newblock \showarticletitle{Graph prototypical networks for few-shot learning on attributed networks}. In \bibinfo{booktitle}{\emph{Proceedings of the 29th ACM International Conference on Information \& Knowledge Management}}. \bibinfo{pages}{295--304}.
\newblock


\bibitem[Du et~al\mbox{.}(2021)]%
        {du2021efficient}
\bibfield{author}{\bibinfo{person}{Jiawei Du}, \bibinfo{person}{Hanshu Yan}, \bibinfo{person}{Jiashi Feng}, \bibinfo{person}{Joey~Tianyi Zhou}, \bibinfo{person}{Liangli Zhen}, \bibinfo{person}{Rick Siow~Mong Goh}, {and} \bibinfo{person}{Vincent~YF Tan}.} \bibinfo{year}{2021}\natexlab{}.
\newblock \showarticletitle{Efficient sharpness-aware minimization for improved training of neural networks}.
\newblock \bibinfo{journal}{\emph{arXiv preprint arXiv:2110.03141}} (\bibinfo{year}{2021}).
\newblock


\bibitem[Finn et~al\mbox{.}(2017)]%
        {finn2017model}
\bibfield{author}{\bibinfo{person}{Chelsea Finn}, \bibinfo{person}{Pieter Abbeel}, {and} \bibinfo{person}{Sergey Levine}.} \bibinfo{year}{2017}\natexlab{}.
\newblock \showarticletitle{Model-agnostic meta-learning for fast adaptation of deep networks}. In \bibinfo{booktitle}{\emph{International conference on machine learning}}. PMLR, \bibinfo{pages}{1126--1135}.
\newblock


\bibitem[Foret et~al\mbox{.}(2020)]%
        {foret2020sharpness}
\bibfield{author}{\bibinfo{person}{Pierre Foret}, \bibinfo{person}{Ariel Kleiner}, \bibinfo{person}{Hossein Mobahi}, {and} \bibinfo{person}{Behnam Neyshabur}.} \bibinfo{year}{2020}\natexlab{}.
\newblock \showarticletitle{Sharpness-aware minimization for efficiently improving generalization}.
\newblock \bibinfo{journal}{\emph{arXiv preprint arXiv:2010.01412}} (\bibinfo{year}{2020}).
\newblock


\bibitem[He et~al\mbox{.}(2016)]%
        {he2016deep}
\bibfield{author}{\bibinfo{person}{Kaiming He}, \bibinfo{person}{Xiangyu Zhang}, \bibinfo{person}{Shaoqing Ren}, {and} \bibinfo{person}{Jian Sun}.} \bibinfo{year}{2016}\natexlab{}.
\newblock \showarticletitle{Deep residual learning for image recognition}. In \bibinfo{booktitle}{\emph{Proceedings of the IEEE conference on computer vision and pattern recognition}}. \bibinfo{pages}{770--778}.
\newblock


\bibitem[Ismail~Fawaz et~al\mbox{.}(2019)]%
        {ismail2019deep}
\bibfield{author}{\bibinfo{person}{Hassan Ismail~Fawaz}, \bibinfo{person}{Germain Forestier}, \bibinfo{person}{Jonathan Weber}, \bibinfo{person}{Lhassane Idoumghar}, {and} \bibinfo{person}{Pierre-Alain Muller}.} \bibinfo{year}{2019}\natexlab{}.
\newblock \showarticletitle{Deep learning for time series classification: a review}.
\newblock \bibinfo{journal}{\emph{Data mining and knowledge discovery}} \bibinfo{volume}{33}, \bibinfo{number}{4} (\bibinfo{year}{2019}), \bibinfo{pages}{917--963}.
\newblock


\bibitem[Karim et~al\mbox{.}(2019)]%
        {karim2019multivariate}
\bibfield{author}{\bibinfo{person}{Fazle Karim}, \bibinfo{person}{Somshubra Majumdar}, \bibinfo{person}{Houshang Darabi}, {and} \bibinfo{person}{Samuel Harford}.} \bibinfo{year}{2019}\natexlab{}.
\newblock \showarticletitle{Multivariate LSTM-FCNs for time series classification}.
\newblock \bibinfo{journal}{\emph{Neural networks}}  \bibinfo{volume}{116} (\bibinfo{year}{2019}), \bibinfo{pages}{237--245}.
\newblock


\bibitem[Keogh and Ratanamahatana(2005)]%
        {keogh2005exact}
\bibfield{author}{\bibinfo{person}{Eamonn Keogh} {and} \bibinfo{person}{Chotirat~Ann Ratanamahatana}.} \bibinfo{year}{2005}\natexlab{}.
\newblock \showarticletitle{Exact indexing of dynamic time warping}.
\newblock \bibinfo{journal}{\emph{Knowledge and information systems}} \bibinfo{volume}{7}, \bibinfo{number}{3} (\bibinfo{year}{2005}), \bibinfo{pages}{358--386}.
\newblock


\bibitem[Kipf and Welling(2016)]%
        {kipf2016semi}
\bibfield{author}{\bibinfo{person}{Thomas~N Kipf} {and} \bibinfo{person}{Max Welling}.} \bibinfo{year}{2016}\natexlab{}.
\newblock \showarticletitle{Semi-supervised classification with graph convolutional networks}.
\newblock \bibinfo{journal}{\emph{arXiv preprint arXiv:1609.02907}} (\bibinfo{year}{2016}).
\newblock


\bibitem[Lai et~al\mbox{.}(2023)]%
        {lai2023context}
\bibfield{author}{\bibinfo{person}{Kwei-Herng Lai}, \bibinfo{person}{Lan Wang}, \bibinfo{person}{Huiyuan Chen}, \bibinfo{person}{Kaixiong Zhou}, \bibinfo{person}{Fei Wang}, \bibinfo{person}{Hao Yang}, {and} \bibinfo{person}{Xia Hu}.} \bibinfo{year}{2023}\natexlab{}.
\newblock \showarticletitle{Context-aware domain adaptation for time series anomaly detection}. In \bibinfo{booktitle}{\emph{Proceedings of the 2023 SIAM International Conference on Data Mining (SDM)}}. SIAM, \bibinfo{pages}{676--684}.
\newblock


\bibitem[Li et~al\mbox{.}(2021)]%
        {li2021shapenet}
\bibfield{author}{\bibinfo{person}{Guozhong Li}, \bibinfo{person}{Byron Choi}, \bibinfo{person}{Jianliang Xu}, \bibinfo{person}{Sourav~S Bhowmick}, \bibinfo{person}{Kwok-Pan Chun}, {and} \bibinfo{person}{Grace Lai-Hung Wong}.} \bibinfo{year}{2021}\natexlab{}.
\newblock \showarticletitle{Shapenet: A shapelet-neural network approach for multivariate time series classification}. In \bibinfo{booktitle}{\emph{Proceedings of the AAAI conference on artificial intelligence}}, Vol.~\bibinfo{volume}{35}. \bibinfo{pages}{8375--8383}.
\newblock


\bibitem[Liu et~al\mbox{.}(2018)]%
        {liu2018time}
\bibfield{author}{\bibinfo{person}{Chien-Liang Liu}, \bibinfo{person}{Wen-Hoar Hsaio}, {and} \bibinfo{person}{Yao-Chung Tu}.} \bibinfo{year}{2018}\natexlab{}.
\newblock \showarticletitle{Time series classification with multivariate convolutional neural network}.
\newblock \bibinfo{journal}{\emph{IEEE Transactions on industrial electronics}} \bibinfo{volume}{66}, \bibinfo{number}{6} (\bibinfo{year}{2018}), \bibinfo{pages}{4788--4797}.
\newblock


\bibitem[Miao et~al\mbox{.}(2024)]%
        {miaorethinking}
\bibfield{author}{\bibinfo{person}{Rui Miao}, \bibinfo{person}{Kaixiong Zhou}, \bibinfo{person}{Yili Wang}, \bibinfo{person}{Ninghao Liu}, \bibinfo{person}{Ying Wang}, {and} \bibinfo{person}{Xin Wang}.} \bibinfo{year}{2024}\natexlab{}.
\newblock \showarticletitle{Rethinking Independent Cross-Entropy Loss For Graph-Structured Data}. In \bibinfo{booktitle}{\emph{Forty-first International Conference on Machine Learning}}.
\newblock


\bibitem[Ruiz et~al\mbox{.}(2021)]%
        {ruiz2021great}
\bibfield{author}{\bibinfo{person}{Alejandro~Pasos Ruiz}, \bibinfo{person}{Michael Flynn}, \bibinfo{person}{James Large}, \bibinfo{person}{Matthew Middlehurst}, {and} \bibinfo{person}{Anthony Bagnall}.} \bibinfo{year}{2021}\natexlab{}.
\newblock \showarticletitle{The great multivariate time series classification bake off: a review and experimental evaluation of recent algorithmic advances}.
\newblock \bibinfo{journal}{\emph{Data Mining and Knowledge Discovery}} \bibinfo{volume}{35}, \bibinfo{number}{2} (\bibinfo{year}{2021}), \bibinfo{pages}{401--449}.
\newblock


\bibitem[Sakoe(1971)]%
        {sakoe1971dynamic}
\bibfield{author}{\bibinfo{person}{Hiroaki Sakoe}.} \bibinfo{year}{1971}\natexlab{}.
\newblock \showarticletitle{Dynamic-programming approach to continuous speech recognition}. In \bibinfo{booktitle}{\emph{1971 Proc. the International Congress of Acoustics, Budapest}}.
\newblock


\bibitem[Snell et~al\mbox{.}(2017)]%
        {snell2017prototypical}
\bibfield{author}{\bibinfo{person}{Jake Snell}, \bibinfo{person}{Kevin Swersky}, {and} \bibinfo{person}{Richard Zemel}.} \bibinfo{year}{2017}\natexlab{}.
\newblock \showarticletitle{Prototypical networks for few-shot learning}.
\newblock \bibinfo{journal}{\emph{Advances in neural information processing systems}}  \bibinfo{volume}{30} (\bibinfo{year}{2017}).
\newblock


\bibitem[Sun et~al\mbox{.}(2024)]%
        {sun2024entropy}
\bibfield{author}{\bibinfo{person}{Mucun Sun}, \bibinfo{person}{Sergio Valdez}, \bibinfo{person}{Juan~M Perez}, \bibinfo{person}{Kevin Garcia}, \bibinfo{person}{Gael Galvan}, \bibinfo{person}{Cesar Cruz}, \bibinfo{person}{Yifeng Gao}, {and} \bibinfo{person}{Li Zhang}.} \bibinfo{year}{2024}\natexlab{}.
\newblock \showarticletitle{Entropy-Infused Deep Learning Loss Function for Capturing Extreme Values in Wind Power Forecasting}. In \bibinfo{booktitle}{\emph{2024 IEEE Green Technologies Conference (GreenTech)}}. IEEE, \bibinfo{pages}{64--68}.
\newblock


\bibitem[Tong et~al\mbox{.}(2022)]%
        {tong2022technology}
\bibfield{author}{\bibinfo{person}{Yuerong Tong}, \bibinfo{person}{Jingyi Liu}, \bibinfo{person}{Lina Yu}, \bibinfo{person}{Liping Zhang}, \bibinfo{person}{Linjun Sun}, \bibinfo{person}{Weijun Li}, \bibinfo{person}{Xin Ning}, \bibinfo{person}{Jian Xu}, \bibinfo{person}{Hong Qin}, {and} \bibinfo{person}{Qiang Cai}.} \bibinfo{year}{2022}\natexlab{}.
\newblock \showarticletitle{Technology investigation on time series classification and prediction}.
\newblock \bibinfo{journal}{\emph{PeerJ Computer Science}}  \bibinfo{volume}{8} (\bibinfo{year}{2022}), \bibinfo{pages}{e982}.
\newblock


\bibitem[Wang et~al\mbox{.}(2024)]%
        {wang2024efficient}
\bibfield{author}{\bibinfo{person}{Yili Wang}, \bibinfo{person}{Kaixiong Zhou}, \bibinfo{person}{Ninghao Liu}, \bibinfo{person}{Ying Wang}, {and} \bibinfo{person}{Xin Wang}.} \bibinfo{year}{2024}\natexlab{}.
\newblock \showarticletitle{Efficient Sharpness-Aware Minimization for Molecular Graph Transformer Models}. In \bibinfo{booktitle}{\emph{The Twelfth International Conference on Learning Representations}}.
\newblock
\urldef\tempurl%
\url{https://openreview.net/forum?id=Od39h4XQ3Y}
\showURL{%
\tempurl}


\bibitem[Wei and Keogh(2006)]%
        {wei2006semi}
\bibfield{author}{\bibinfo{person}{Li Wei} {and} \bibinfo{person}{Eamonn Keogh}.} \bibinfo{year}{2006}\natexlab{}.
\newblock \showarticletitle{Semi-supervised time series classification}. In \bibinfo{booktitle}{\emph{Proceedings of the 12th ACM SIGKDD international conference on Knowledge discovery and data mining}}. \bibinfo{pages}{748--753}.
\newblock


\bibitem[Xi et~al\mbox{.}(2023)]%
        {xi2023lb}
\bibfield{author}{\bibinfo{person}{Wenjie Xi}, \bibinfo{person}{Arnav Jain}, \bibinfo{person}{Li Zhang}, {and} \bibinfo{person}{Jessica Lin}.} \bibinfo{year}{2023}\natexlab{}.
\newblock \showarticletitle{Lb-simtsc: An efficient similarity-aware graph neural network for semi-supervised time series classification}.
\newblock \bibinfo{journal}{\emph{arXiv preprint arXiv:2301.04838}} (\bibinfo{year}{2023}).
\newblock


\bibitem[Xi et~al\mbox{.}(2024)]%
        {xi2024efficient}
\bibfield{author}{\bibinfo{person}{Wenjie Xi}, \bibinfo{person}{Arnav Jain}, \bibinfo{person}{Li Zhang}, {and} \bibinfo{person}{Jessica Lin}.} \bibinfo{year}{2024}\natexlab{}.
\newblock \showarticletitle{Efficient and Accurate Similarity-Aware Graph Neural Network for Semi-supervised Time Series Classification}. In \bibinfo{booktitle}{\emph{Pacific-Asia Conference on Knowledge Discovery and Data Mining}}. Springer, \bibinfo{pages}{276--287}.
\newblock


\bibitem[Xue et~al\mbox{.}(2021)]%
        {xue2021cap}
\bibfield{author}{\bibinfo{person}{Haotian Xue}, \bibinfo{person}{Kaixiong Zhou}, \bibinfo{person}{Tianlong Chen}, \bibinfo{person}{Kai Guo}, \bibinfo{person}{Xia Hu}, \bibinfo{person}{Yi Chang}, {and} \bibinfo{person}{Xin Wang}.} \bibinfo{year}{2021}\natexlab{}.
\newblock \showarticletitle{CAP: Co-Adversarial Perturbation on Weights and Features for Improving Generalization of Graph Neural Networks}.
\newblock \bibinfo{journal}{\emph{arXiv preprint arXiv:2110.14855}} (\bibinfo{year}{2021}).
\newblock


\bibitem[Zha et~al\mbox{.}(2022)]%
        {zha2022towards}
\bibfield{author}{\bibinfo{person}{Daochen Zha}, \bibinfo{person}{Kwei-Herng Lai}, \bibinfo{person}{Kaixiong Zhou}, {and} \bibinfo{person}{Xia Hu}.} \bibinfo{year}{2022}\natexlab{}.
\newblock \showarticletitle{Towards similarity-aware time-series classification}. In \bibinfo{booktitle}{\emph{Proceedings of the 2022 SIAM International Conference on Data Mining (SDM)}}. SIAM, \bibinfo{pages}{199--207}.
\newblock


\bibitem[Zhang et~al\mbox{.}(2017)]%
        {zhang2017mixup}
\bibfield{author}{\bibinfo{person}{Hongyi Zhang}, \bibinfo{person}{Moustapha Cisse}, \bibinfo{person}{Yann~N Dauphin}, {and} \bibinfo{person}{David Lopez-Paz}.} \bibinfo{year}{2017}\natexlab{}.
\newblock \showarticletitle{mixup: Beyond empirical risk minimization}.
\newblock \bibinfo{journal}{\emph{arXiv preprint arXiv:1710.09412}} (\bibinfo{year}{2017}).
\newblock


\bibitem[Zhang et~al\mbox{.}(2022)]%
        {zhang2022joint}
\bibfield{author}{\bibinfo{person}{Li Zhang}, \bibinfo{person}{Nital Patel}, \bibinfo{person}{Xiuqi Li}, {and} \bibinfo{person}{Jessica Lin}.} \bibinfo{year}{2022}\natexlab{}.
\newblock \showarticletitle{Joint time series chain: Detecting unusual evolving trend across time series}. In \bibinfo{booktitle}{\emph{Proceedings of the 2022 SIAM International Conference on Data Mining (SDM)}}. SIAM, \bibinfo{pages}{208--216}.
\newblock


\bibitem[Zhang et~al\mbox{.}(2020)]%
        {zhang2020tapnet}
\bibfield{author}{\bibinfo{person}{Xuchao Zhang}, \bibinfo{person}{Yifeng Gao}, \bibinfo{person}{Jessica Lin}, {and} \bibinfo{person}{Chang-Tien Lu}.} \bibinfo{year}{2020}\natexlab{}.
\newblock \showarticletitle{Tapnet: Multivariate time series classification with attentional prototypical network}. In \bibinfo{booktitle}{\emph{Proceedings of the AAAI conference on artificial intelligence}}, Vol.~\bibinfo{volume}{34}. \bibinfo{pages}{6845--6852}.
\newblock


\bibitem[Zhou et~al\mbox{.}(2020)]%
        {zhou2020towards}
\bibfield{author}{\bibinfo{person}{Kaixiong Zhou}, \bibinfo{person}{Xiao Huang}, \bibinfo{person}{Yuening Li}, \bibinfo{person}{Daochen Zha}, \bibinfo{person}{Rui Chen}, {and} \bibinfo{person}{Xia Hu}.} \bibinfo{year}{2020}\natexlab{}.
\newblock \showarticletitle{Towards deeper graph neural networks with differentiable group normalization}.
\newblock \bibinfo{journal}{\emph{Advances in neural information processing systems}}  \bibinfo{volume}{33} (\bibinfo{year}{2020}), \bibinfo{pages}{4917--4928}.
\newblock


\bibitem[Zhou et~al\mbox{.}(2021)]%
        {zhou2021dirichlet}
\bibfield{author}{\bibinfo{person}{Kaixiong Zhou}, \bibinfo{person}{Xiao Huang}, \bibinfo{person}{Daochen Zha}, \bibinfo{person}{Rui Chen}, \bibinfo{person}{Li Li}, \bibinfo{person}{Soo-Hyun Choi}, {and} \bibinfo{person}{Xia Hu}.} \bibinfo{year}{2021}\natexlab{}.
\newblock \showarticletitle{Dirichlet energy constrained learning for deep graph neural networks}.
\newblock \bibinfo{journal}{\emph{Advances in Neural Information Processing Systems}}  \bibinfo{volume}{34} (\bibinfo{year}{2021}), \bibinfo{pages}{21834--21846}.
\newblock


\end{thebibliography}
\end{document}